\documentclass[runningheads]{llncs}
\usepackage{graphicx}
\usepackage{tikz}
\usepackage{comment}
\usepackage{amsmath,amssymb} % define this before the line numbering.
\usepackage{color}
\usepackage{array}
\usepackage{multirow}
\usepackage{pifont}

\usepackage{hyperref}
\hypersetup{
	colorlinks=true,
	linkcolor=blue,
	filecolor=magenta,      
	urlcolor=magenta,
	pdftitle={Overleaf Example},
	pdfpagemode=FullScreen,
}

\begin{document}
	% \linenumbers
	\pagestyle{headings}
	\mainmatter
	\def\ECCVSubNumber{5940}  % Insert your submission number here
	
	\title{SPSN: Superpixel Prototype Sampling Network for RGB-D Salient Object Detection} % Replace with your title
	
	\begin{comment}
		\titlerunning{ECCV-20 submission ID \ECCVSubNumber} 
		\authorrunning{ECCV-20 submission ID \ECCVSubNumber} 
		\author{Anonymous ECCV submission}
		\institute{Paper ID \ECCVSubNumber}
	\end{comment}
	%******************
	
	%\begin{comment}
	\titlerunning{SPSN: Superpixel Prototype Sampling Network for RGB-D SOD}
	\author{Minhyeok Lee\thanks{These authors contributed equally} \and
		Chaewon Park\inst{\star} \and
		Suhwan Cho \and
		Sangyoun Lee}
	\authorrunning{M. Lee et al.}
	\institute{Yonsei University, Seoul, Korea \\
		\email{\{hydragon516, chaewon28, chosuhwan, syleee\}@yonsei.ac.kr}}
	%\end{comment}
	%******************
	\maketitle
	
	\begin{abstract}
		RGB-D salient object detection (SOD) has been in the spotlight recently because it is an important preprocessing operation for various vision tasks. However, despite advances in deep learning-based methods, RGB-D SOD is still challenging due to the large domain gap between an RGB image and the depth map and low-quality depth maps. To solve this problem, we propose a novel superpixel prototype sampling network (SPSN) architecture. The proposed model splits the input RGB image and depth map into component superpixels to generate component prototypes. We design a prototype sampling network so that the network only samples prototypes corresponding to salient objects. In addition, we propose a reliance selection module to recognize the quality of each RGB and depth feature map and adaptively weight them in proportion to their reliability. The proposed method makes the model robust to inconsistencies between RGB images and depth maps and eliminates the influence of non-salient objects. Our method is evaluated on five popular datasets, achieving state-of-the-art performance. We prove the effectiveness of the proposed method through comparative experiments. Code and models are available at \url{https://github.com/Hydragon516/SPSN}.
		
	\end{abstract}

	\section{Introduction}
	The salient object detection (SOD) task detects and segments objects that visually attract the most human interest from a single image or video. The SOD task is a useful preprocessing operation for various computer vision tasks such as few-shot learning, weakly-supervised semantic segmentation, object recognition, tracking, and image parsing. However, despite recent advances in deep learning, it is still challenging due to camouflaged objects, extreme lighting conditions, and scenes containing multiple objects with complex shapes. To potentially improve performance for such difficult scenes, RGB-D SOD, using an additional depth map, has recently been in the spotlight.
	
	\begin{figure}[t]
		\setlength{\belowcaptionskip}{-24pt}
		\begin{center}
			\includegraphics[width=\linewidth]{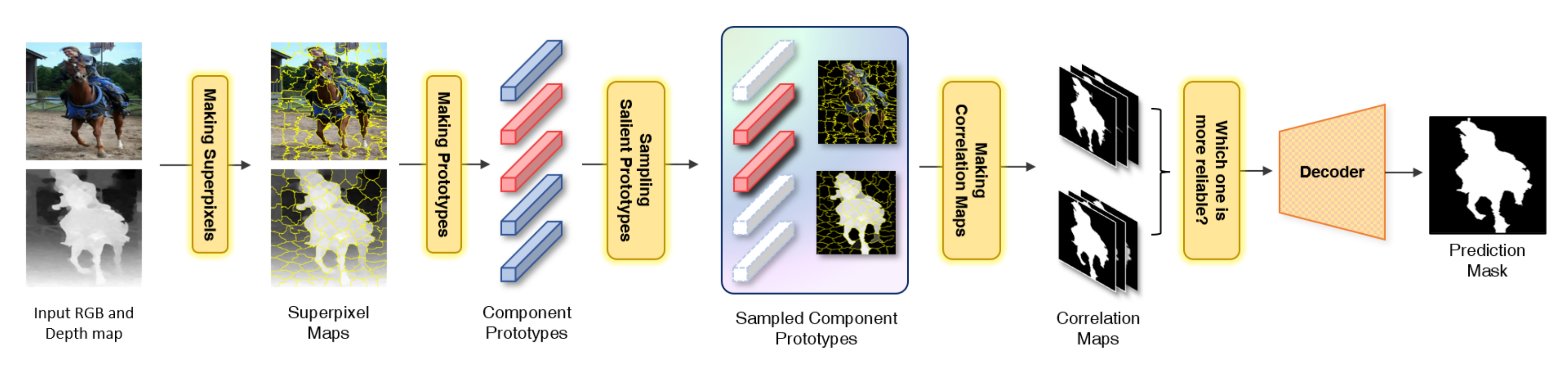}
			\caption{The overall flow of the proposed model. Our model generates and samples component prototypes from superpixel maps. It also compares the reliability of correlation maps created from component prototypes to generate the predicted mask.}
			\label{fig:intro}
		\end{center}
	\end{figure}
	
	Recent deep learning-based studies~\cite{chen2017m,chen2018progressively,chen2018attention,zhao2019contrast,fu2020jl} achieve significant RGB-D SOD performance by fusing RGB information and additional depth information. However, due to the large domain gap between an RGB image containing rich detail information and a depth image containing geometric information, previous works~\cite{chen2018progressively,chen2019three,qu2017rgbd,song2017depth,fan2020rethinking,piao2019depth} focus on the process of effectively fusing these two pieces of information. These methods show that they can effectively extract feature information about salient objects from RGB images and depth maps, but they have two major limitations.
	
	First, they perform inconsistently due to mismatches between the RGB image and the depth map. For example, in the case of a picture hung on the wall, the depth map lacks saliency information compared to the RGB image due to the picture’s thinness. Furthermore, the RGB image contains complex texture information about the background scene despite the particularly monotonous depth map background. This unnecessary additional information acts as noise in the network and makes it difficult to generate an accurate saliency mask. This often causes conventional methods to fail in challenging scenes involving complex background structures and multiple foreground objects. 
	
	Second, the quality of depth maps is inconsistent due to the limitations of the depth sensor. Some studies~\cite{ji2021calibrated,sun2021deep} suggest additional processes for depth map refinement to solve this problem. Although these methods can improve the consistency of low-quality depth maps, they are inefficient due to the additional network or computational costs.
	
	To solve the problems described above, we propose a novel superpixel prototype sampling network (SPSN) architecture. Fig.~\ref{fig:intro} shows the overall SPSN process. First, we note that RGB images and depth maps provide different kinds of information and can complement each other. RGB images have various detail and texture information in the foreground and background, which provides rich context information to the network as it passes through the encoder. In comparison, the depth map lacks detailed information, but it is more robust than an RGB image in extreme lighting conditions. For a preprocessing operation to effectively fuse and complement the advantages of an RGB image and depth map, we use the simple linear iterative clustering (SLIC) algorithm~\cite{achanta2012slic} to segment the RGB image and depth map into superpixel components. Moreover, we propose a prototype sampling network module (PSNM) to solve the inconsistency problem between RGB images and depth maps and extract salient object features effectively. We generate component prototypes from superpixel components, inspired by prototype learning, which is used extensively in few-shot segmentation tasks~\cite{wang2019panet,li2021adaptive,liu2020part}. PSNM, composed of transformers and graph convolutional layers, is trained to selectively sample only prototypes corresponding to salient objects among component prototypes. Therefore, the proposed method improves performance by minimizing the influence of the background and extracting consistent salient features from RGB images and depth maps. Furthermore, for the network to be flexible enough to handle low-quality depth maps, we propose a reliance selection module (RSM). The RSM is trained to evaluate the quality of the features generated from the RGB component prototypes and depth component prototypes. As a result, the RSM adaptively changes the RGB image and depth map dependence of the network. In other words, the proposed model minimizes performance degradation in situations such as low-quality depth maps and low-light RGB images and effectively creates a saliency mask.

	The experimental results over five benchmark datasets show that our model significantly outperforms previous state-of-the-art approaches. Finally, we demonstrate the validity of our method through various ablation studies.
	
	\section{Related Work}
	\label{sec:related}
	\subsection{RGB-D SOD}
	Recent RGB-based SOD methods~\cite{zhao2015saliency,wang2019salient,zhao2019egnet,park2021saliency,fan2018salient} have demonstrated outstanding performance. However, they are still challenged by insufficient information to express the complex characteristics of scenes with multiple objects, transparent objects, ambiguous borders between the foreground and background, and extreme light conditions. Meanwhile, owing to the development of various consumer-grade depth cameras, additional depth cues of abundant structural and geometrical information have been enabled for SOD studies. Therefore, RGB-D SOD has gained significant attention and has been widely studied to supplement the limits of RGB-based methods on the scenarios mentioned above. 
	
	\subsection{Traditional RGB-D Methods}
	Traditional RGB-D SOD algorithms~\cite{desingh2013depth,cong2016saliency,cheng2014depth,cong2019going,ju2014depth,guo2016salient,lang2012depth} focused on utilizing various hand-crafted features, such as contrast, center or boundary prior, and center-surround difference. 
	Lang~{\textit{et al.}}~\cite{lang2012depth} introduced the depth prior by modeling the relationship between depth and saliency with a mixture of Gaussians. 
	Additionally, Cheng~{\textit{et al.}}~\cite{cheng2014depth} grouped the pixels in the input image with k-means clustering and obtained three saliency cues—color contrast, depth contrast and spatial bias—from each cluster to generate saliency maps.
	Moreover, Ju~{\textit{et al.}}~\cite{ju2014depth} proposed an anisotropic center-surround difference based on the assumption that salient objects tend to stand out from the surroundings. 
	Because these methods rely heavily on hand-crafted features of relatively limited information, their performance deteriorates in complex scenes.
	
	\subsection{Deep Learning-Based RGB-D Methods}
	Existing deep learning-based RGB-D SOD methods focus more on fusing the complementary features extracted from the RGB and depth channels because the domain gap represented by each channel is significant. The merging strategies can be grouped into three categories based on when the fusion takes place: early fusion~\cite{qu2017rgbd,song2017depth}, middle fusion~\cite{chen2018progressively,chen2019three}, and late fusion~\cite{fan2020rethinking,piao2019depth}. 
	Early fusion methods concatenate the RGB image and depth image at the earliest stage and regard the integrated four-channel matrix as a single input. For example, Qu~{\textit{et al.}}~\cite{qu2017rgbd} introduced this method by generating hand-crafted feature vectors from each RGB-D pair, which were fed as input to a CNN-based model. 
	Middle fusion methods fuse the two different feature maps extracted from individual networks. For example, Chen~{\textit{et al.}}~\cite{chen2018progressively} suggested a two-stream complementary-aware network in which the features from the same stages of each modality are fused with the help of a complementary-aware fusion block.
	Finally, late fusion methods produce individual saliency prediction maps from both the RGB and depth channels, and the two predicted maps are merged by a post-processing operation such as pixel-wise summation and multiplication. For example, Piao~{\textit{et al.}}~\cite{piao2019depth} proposed a depth-induced multiscale recurrent attention network to extract the features from an RGB image and depth image individually and designed depth refinement blocks for integration.
	
	However, these methods neglect the problem of mismatches between the two modalities. For example, in some scenarios such as a thin calendar hung on the wall, the RGB image more accurately discriminates the salient object and the background, whereas all the pixel values are similar to each other in the depth image. To deal with this problem, several studies have proposed methods to enhance such unreliable input data by utilizing hand-crafted techniques to improve the accuracy. Zhao~{\textit{et al.}}~\cite{zhao2019contrast} suggested a contrast prior loss to increase the color difference between the foreground and background of the depth input. Ji~{\textit{et al.}}~\cite{ji2021calibrated} proposed an effective depth calibration strategy that corrects the latent bias of the raw depth maps. Furthermore, Zhang~{\textit{et al.}}~\cite{zhang2020uc} presented a depth correction network to decrease the noise in unreliable depth data, assuming the object boundaries in the depth map align with those in the RGB map.
	
	\begin{figure}[t]
		\setlength{\belowcaptionskip}{-24pt}
		\begin{center}
			\includegraphics[width=\linewidth]{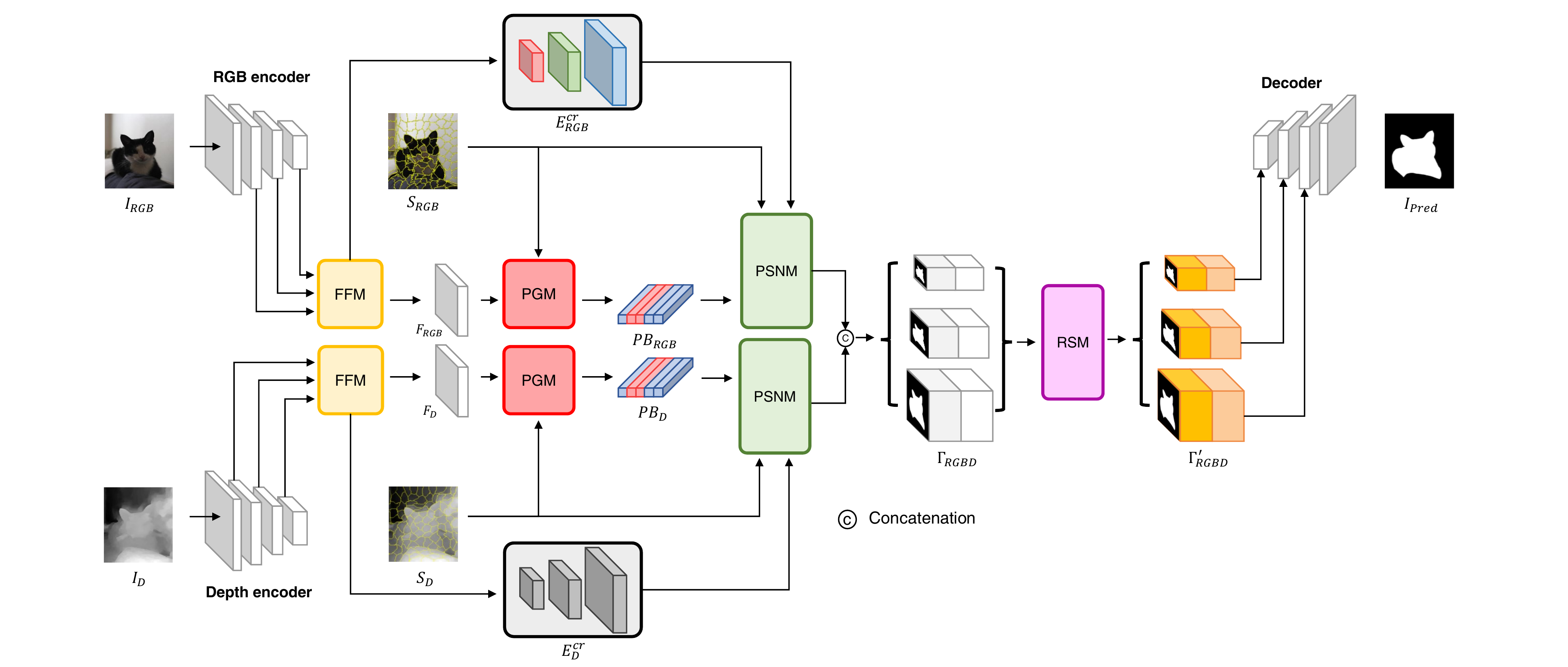}
			\caption{Overall architecture of the superpixel prototype sampling network (SPSN). The proposed network has one RGB encoder and one depth encoder. Our model consists of a feature fusion model (FFM) for effective fusion of encoder features, prototype generating module (PGM) for prototype extraction, prototype sampling network module (PSNM) for prototype sampling, and reliance selection module (RSM) for reliability selection of RGB and depth features.}
			\label{fig:model}
		\end{center}
	\end{figure}
	
	\section{Proposed Method}
	\subsection{Overview}
	Fig.~\ref{fig:model} shows the overall architecture of the proposed SPSN. The proposed model uses an RGB image $\mathbf{I_{RGB}}$, depth map $\mathbf{I_D}$, and their superpixel maps $\mathbf{S_{RGB}}$, $\mathbf{S_{D}}$ as inputs. Our model is composed primarily of four parts: the feature fusion module (FFM), prototype generating module (PGM), prototype sampling network module (PSNM), and reliance selection module (RSM). The SPSN also has two encoders for RGB images and depth maps and one decoder.
	
	\subsection{Feature Fusion Module}
	As shown in Fig.~\ref{fig:module1&2} (a), FFM fuses multiscale features from the encoder. We extract three features $ \mathbf{E_1}  \in \mathbb{R} ^ {C_{ \left ( 1/8 \right )} \times \frac{ H } { 8 } \times \frac{ W } { 8 }}$, $ \mathbf{E_2}  \in \mathbb{R} ^ {C_{ \left ( 1/16 \right )} \times \frac{ H } { 16 } \times \frac{ W } { 16 }}$, and $ \mathbf{E_3} \in \mathbb{R} ^ {C_{ \left ( 1/32 \right )} \times \frac{ H } { 32 } \times \frac{ W } { 32 }}$ from the encoder, where $H$ and $W$ are the height and width of the input image, respectively, and $C_{ \left ( 1/8 \right )}$, $C_{ \left ( 1/16 \right )}$ , and $C_{ \left ( 1/32 \right )}$ are the number of channels of the multiscale encoder feature. Because the architectures of the RGB encoder and depth encoder are identical, the size of the extracted features is the same. The FFM consists of a $1 \times 1$ convolution layer and upsampling layers, integrating the multiscale features of the encoder and extracting the global contextual information through atrous spatial pyramid pooling (ASPP)~\cite{chen2017deeplab} layer. As a result, The FFM generates RGB fusion feature $\mathbf{F_{RGB}} \in \mathbb{R} ^ {128 \times \frac{ H } { 8 } \times \frac{ W } { 8 }}$ and depth fusion feature $\mathbf{F_D} \in \mathbb{R} ^ {128 \times \frac{ H } { 8 } \times \frac{ W } { 8 }}$, as shown in Fig.~\ref{fig:model}. In addition, the channel-reduced encoder feature $\mathbf{E^{cr}}$ after the $1 \times 1$ convolution layer is used as the input for PSNM.
	
	\subsection{Prototype Generating Module}
\begin{figure}[t]
		\setlength{\belowcaptionskip}{-24pt}
		\begin{center}
			\includegraphics[width=0.9\linewidth]{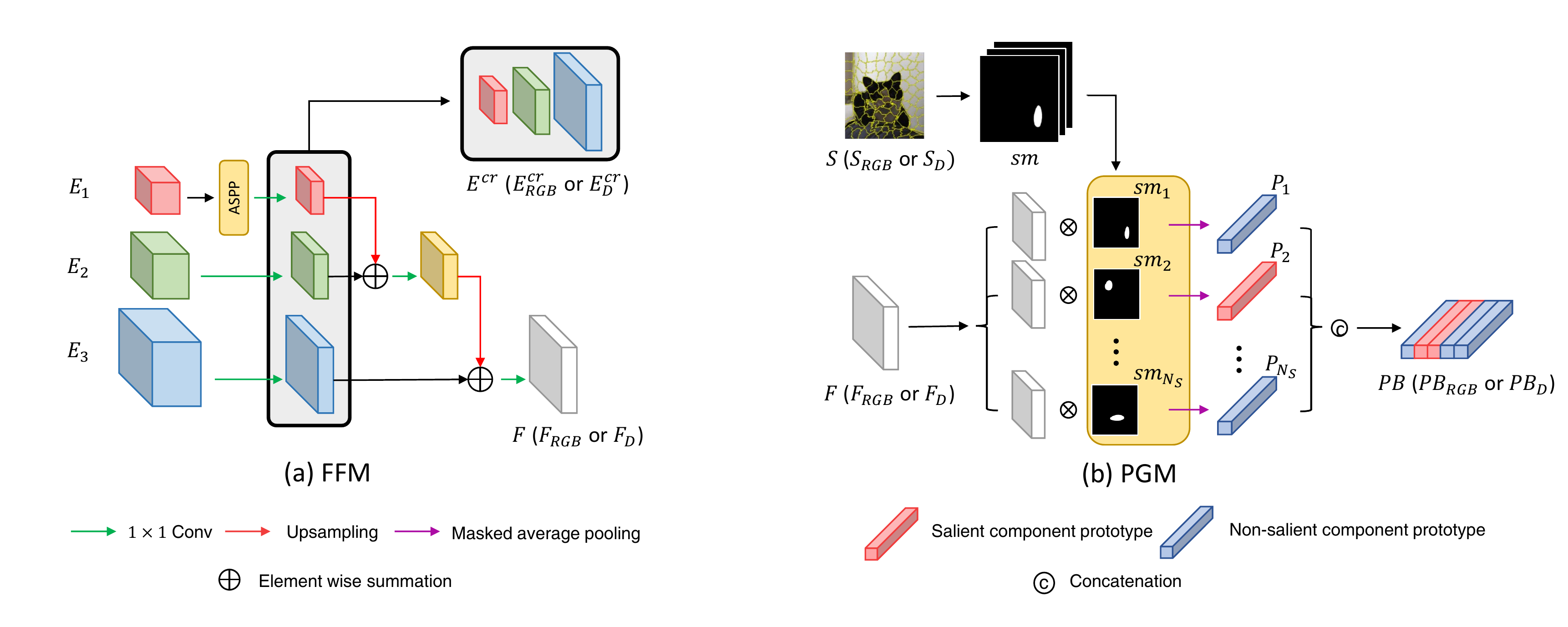}
			\caption{Structure of (a) FFM and (b) PGM. The FFM fuses the multiscale features of the encoder. The PGM generates a prototype block from the superpixel mask $\mathbf{sm}$ and the fusion feature $F$ generated from the FFM.}
			\label{fig:module1&2}
		\end{center}
	\end{figure}

	The PGM aims to generate component prototypes from fusion features $\mathbf{F_{RGB}}$ and $\mathbf{F_D}$, obtained from the FFM. As shown in Fig.~\ref{fig:module1&2} (b), we first create a superpixel map $\mathbf{S}$ from each RGB image $\mathbf{I_{RGB}}$ and depth map $\mathbf{I_D}$ using the SLIC algorithm~\cite{achanta2012slic}. Next, we create a superpixel mask group $\mathbf{sm}$ where each channel is a binary mask for each superpixel. Therefore, if the number of superpixels is $N_S$, the size of $\mathbf{sm}$ is $N_S \times H \times W$. $\mathbf{sm}$ is down-sampled to the same size as the fusion feature $\mathbf{F}$ (\textit{i}.\textit{e}. $\mathbf{F_{RGB}}$ or $\mathbf{F_D}$) generated by the FFM, so the size of the superpixel mask $\mathbf{sm_i}$ constituting each channel of $\mathbf{sm}$ is $1 \times \frac{ H } { 8 } \times \frac{ W } { 8 }$, where $i = 1, 2, ..., N_S$. Like the prototype learning of few-shot segmentation tasks~\cite{wang2019panet,li2021adaptive,liu2020part}, the PGM creates a component prototype from each superpixel mask. Thus, the prototype $\mathbf{P _ { i }}$ generated by $\mathbf{sm_i}$ is defined as $\mathbf{P _ { i }} =MAP \left ( \mathbf{F}, \mathbf{sm_i} \right )$, where $MAP\left( . \right)$ is the masked average pooling operator. Finally, we define a prototype block $\mathbf{PB} \in \mathbb{R} ^ {N_S \times 128}$ as a concatenation of the component prototypes generated. As shown in Fig.~\ref{fig:module1&2} (b), we define prototypes created from superpixel masks on salient objects as salient component prototypes, and prototypes created from superpixel masks at other locations as non-salient component prototypes.
	
	\subsection{Prototype Sampling Network Module}
	\label{PSNM}
\begin{figure}[t]
		\setlength{\belowcaptionskip}{-24pt}
		\begin{center}
			\includegraphics[width=\linewidth]{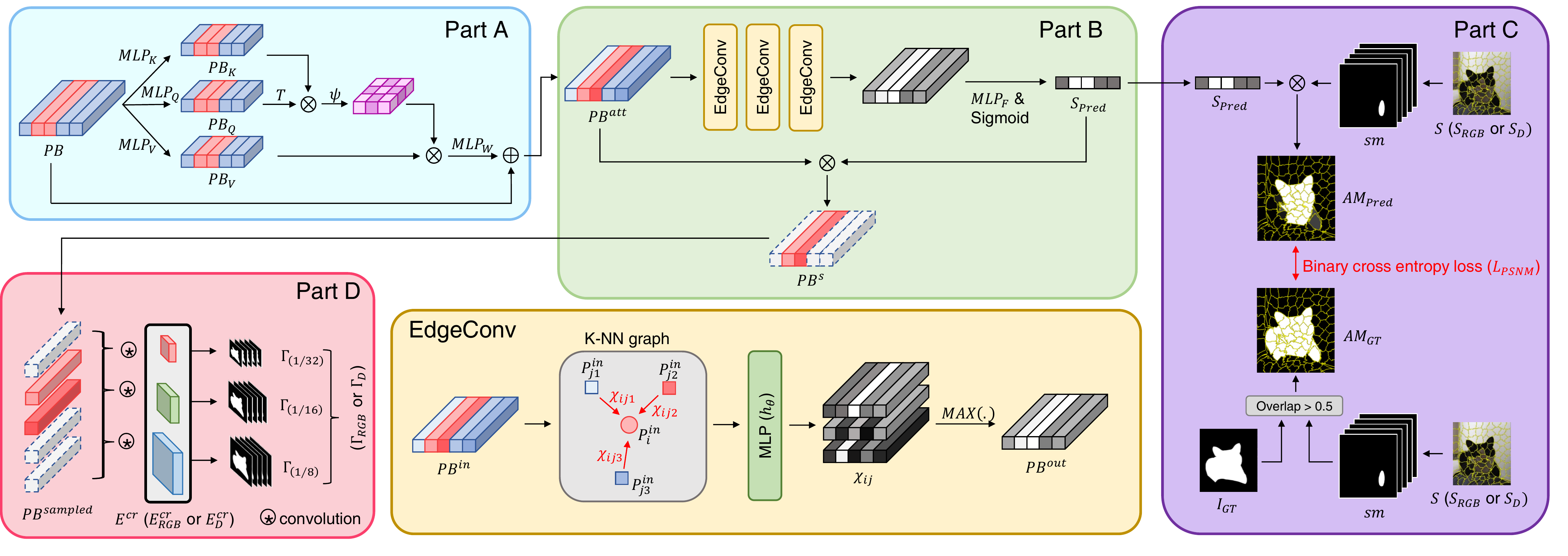}
			\caption{Structure of the PSNM, composed primarily of four subparts. The PSNM selectively samples only prototypes corresponding to salient objects.}
			\label{fig:module3}
		\end{center}
	\end{figure}

	The PSNM aims to sample only the salient component prototypes from all the component prototypes $\mathbf{P _ { i }}$ created from the RGB images and depth maps. Therefore, the PSNM should focus on correlations between $\mathbf{P _ { i }}$ that contain consistent characteristics for salient objects, and it must be able to distinguish them from inconsistent background components. Fig.~\ref{fig:module3} shows the structure of the proposed PSNM, which consists of Parts A, B, C, and D.
	
	\noindent
	\textbf{Part A.} Part A is a transformer module with multi-head attention to enhance the correlation between $\mathbf{P _ { i }}$. Inspired by the previous key, query, and value-based multi-head attention method~\cite{wang2018non,fu2019dual,zhang2019self}, we first generate $\mathbf{PB_K} \in \mathbb{R} ^ {N_S \times 64}$, $\mathbf{PB_Q} \in \mathbb{R} ^ {N_S \times 64}$, and $\mathbf{PB_V} \in \mathbb{R} ^ {N_S \times 64}$ from the prototype block $\mathbf{PB}$ using MLP blocks $MLP _ { K }$, $MLP _ { Q }$, and $MLP _ { V }$. By the MLP block, the length of prototypes is reduced by half, with each prototype block defined as $\mathbf{PB _ { K }} =MLP _ { K } \left ( \mathbf{PB} \right )$, $\mathbf{PB _ { Q }} =MLP _ { Q } \left ( \mathbf{PB} \right )$, and $\mathbf{PB _ { V }} =MLP _ { V } \left ( \mathbf{PB} \right )$. The Part A output $\mathbf{PB ^ { att }} \in \mathbb{R} ^ {N_S \times 128}$ is defined by the following equation:
	
	\begin{equation}
		\mathbf{PB ^ { att }} = \mathbf{PB} + MLP _ { W } \left ( \psi \left ( \frac{\mathbf{PB_Q} \cdot \left ( \mathbf{PB_K} \right ) ^ { T } } { \sqrt { d } } \right ) \cdot \mathbf{PB_V} \right ),
	\end{equation}
	
	\noindent
	where $\left ( . \right ) ^ { T }$ and $\psi \left ( . \right )$ are the transpose and softmax operators, respectively. Furthermore, $d = 64$, the length of $\mathbf{PB_K}$, $\mathbf{PB_Q}$, and $\mathbf{PB_V}$. In addition, $MLP _ { W }$ is an MLP block that increases the length of the reduced prototype to the length of the original.
	
	\noindent
	\textbf{Part B.} Part B is a network for sampling salient component prototypes from $\mathbf{PB ^ { att }}$ with enhanced correlation between component prototypes. Since the component prototype is the result of masked average pooling of the encoder features, it is a one-dimensional vector representing each component feature. These feature shapes are similar to the embedded point features of 3D point cloud networks~\cite{qi2017pointnet,qi2017pointnet++,wang2019dynamic}. Therefore, we propose a graph convolution network based on the feature distance between the prototypes, inspired by EdgeConv~\cite{wang2019dynamic}, used in 3D point cloud networks. As shown in Part B of Fig.~\ref{fig:module3}, the proposed module consists of three EdgeConv layers and one MLP block. First, we define the input component prototype block of EdgeConv as $\mathbf{PB^{in}} \in \mathbb{R} ^ {N_S \times 128}$ containing prototypes $\mathbf{P_{1}^{in}}, \mathbf{P_{2}^{in}}, ..., \mathbf{P_{N_S}^{in}}$. Then, EdgeConv uses the k-nearest-neighbor (k-NN) algorithm to create a graph between a target prototype $\mathbf{P_i^{in}}$ and $a_k$ prototypes $\mathbf{P_{j1}^{in}}, \mathbf{P_{j2}^{in}}, ..., \mathbf{P_{ja_k}^{in}}$ that are most close to each other in the feature space. Next, as shown in Fig.~\ref{fig:module3}, EdgeConv extracts edge features $\mathbf{\chi _ { ijx }}$ between each node generated in the graph, where $x = 1, 2, ..., a_k$. The edge features $\mathbf{\chi _ { ijx }}$ are defined as follows:
	
	\begin{equation}
		\mathbf{\chi _ { ijx }} =h _ { \theta } \left ( \mathbf{P _ { i } ^ {in}}, \mathbf{P _ { jx } ^ {in}} - \mathbf{P _ { i } ^ {in}} \right ),
	\end{equation}
	
	\noindent
	where $h _ { \theta } :\mathbb R ^ { c _ { e } } \times \mathbb R ^ { c _ { e } } \rightarrow \mathbb R ^ { c _ { e } }$ is a nonlinear function with a set of learnable parameters $\theta$, and $c_e = 128$. Therefore, as shown in Fig.~\ref{fig:module3}, a total of $a_k$ $\mathbf{\chi _ { ijx }}$ are generated, so the size of $\mathbf{\chi _ { ij }}$ is $a_k \times N_S \times 128$. This process is equivalent to generating dynamic graphs proposed by~\cite{wang2019dynamic}. The final output $\mathbf{PB^{out}} \in \mathbb{R} ^ {N_S \times 128}$ of the EdgeConv layer is defined as $\mathbf{PB^{out}} = MAX \left( \mathbf{\chi _ { ij }} \right)$, where $MAX \left(. \right)$ is the channel-wise symmetric aggregation operator max pooling, according to~\cite{wang2019dynamic}. The symmetric aggregation operator makes the network independent of the prototype order. As a result, Part B generates a prototype sampler vector $\mathbf{S_{pred}} \in \mathbb{R} ^ {N_S}$, defined as follows:
	
	\begin{equation}
		\mathbf{S _ { pred }} =Sigmoid \left ( MLP _ { F } \left (\mathbf{PB ^ { out }} \right ) \right ),
	\end{equation}
	
	\noindent
	where $Sigmoid \left ( . \right )$ is the sigmoid operator and $MLP _ { F }$ is an MLP block that reduces the length of the prototype block. Therefore, as shown in Fig.~\ref{fig:module3}, $\mathbf{S _ { pred }}$ has values between 0 and 1, and we multiply $\mathbf{S _ { pred }}$ by $\mathbf{PB ^ { att }}$ to create a $\mathbf{PB ^ { s }}$, where only the salient object is sampled.
	
	\noindent
	\textbf{Part C.} Part C is an auxiliary module for training the network of Part B. It is used only in the training phase and is removed in the testing phase. As shown in Part C in Fig.~\ref{fig:module3}, we compute the channel-wise sum of the multiplication of $\mathbf{S_{Pred}}$ and $\mathbf{m_S}$ to generate the auxiliary prediction superpixel map $\mathbf{AM_{Pred}} \in \mathbb{R} ^ {H \times W}$. In other words, $\mathbf{AM_{Pred}}$ is the set of superpixel masks sampled by $\mathbf{S_{Pred}}$. We also generate an auxiliary ground truth superpixel map $\mathbf{AM_{GT}} \in \mathbb{R} ^ {H \times W}$ from the $\mathbf{sm}$ and the ground truth salient object mask $\mathbf{I_{GT}}$. $\mathbf{AM_{GT}}$ is the channel-wise sum of $\mathbf{sm}$ satisfying $\frac{ \sum \left ( \mathbf{m _ { S _ { k } }} \times \mathbf{I _ { GT }} \right ) } { \sum \left ( \mathbf{m _ { S _ { k } }} \right ) } >0.5$, where $\sum \left( . \right)$ is the sum of all pixel values. Therefore, as shown in Part C of Fig.~\ref{fig:module3}, $\mathbf{AM_{GT}}$ is similar to $\mathbf{I_{GT}}$. We use the binary cross-entropy loss between $\mathbf{AM_{Pred}}$ and $\mathbf{AM_{GT}}$ as an objective function so that Part B can learn to sample only the salient object prototypes.
	
	\noindent
	\textbf{Part D.} Part D generates correlation features for the salient objects from $\mathbf{PB^{s}}$ and $\mathbf{E^{cr}}$. We treat each of the prototype blocks $\mathbf{PB^{s}_1}, \mathbf{PB^{s}_2}, ... , \mathbf{PB^{s}_{N_S}}$ as a $1 \times 1$ convolution kernel and perform convolution with $\mathbf{E^{cr}}$. As shown in Part D of Fig.~\ref{fig:module3}, the correlation maps $\mathbf{\Gamma_{\left( 1/32 \right)}} \in \mathbb{R} ^ {N_S \times \frac{ H } { 32 } \times \frac{ W } { 32 }}$, $\mathbf{\Gamma_{\left( 1/16 \right)}} \in \mathbb{R} ^ {N_S \times \frac{ H } { 16 } \times \frac{ W } { 16 }}$, and $\mathbf{\Gamma_{\left( 1/8 \right)}} \in \mathbb{R} ^ {N_S \times \frac{ H } { 8 } \times \frac{ W } { 8 }}$ are generated by channel-wise concatenation of convolution results by multiscale $\mathbf{E^{cr}}$s and each $1 \times 1$ kernel. This process makes it possible to exclude non-salient object features and generate correlation maps for salient objects.
	
	\begin{figure}[t]
		\setlength{\belowcaptionskip}{-24pt}
		\begin{center}
			\includegraphics[width=0.9\linewidth]{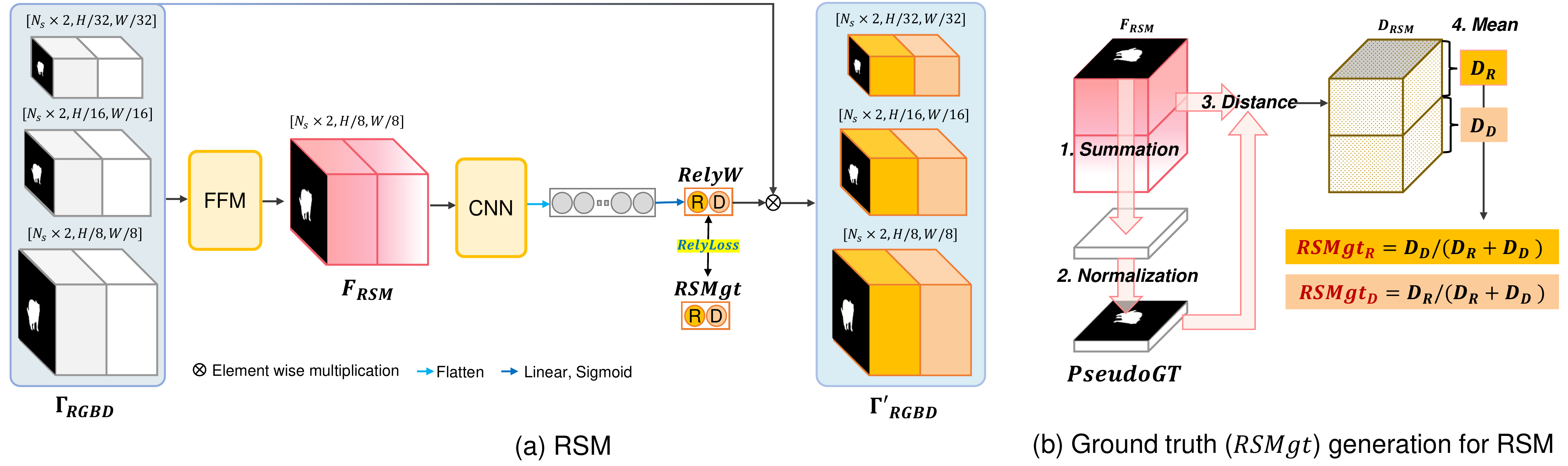}
			\caption{(a) Structure of the RSM and (b) ground truth generating process for the RSM. The RSM aims to discriminate the reliability of each RGB feature and depth feature to adaptively balance the contribution of the two when generating the final saliency map.}
			\label{fig:RSM}
		\end{center}
	\end{figure} 
	
	\subsection{Reliance Selection Module}
	\label{RSM}
	As previously mentioned, reliable modality varies depending on the characteristics of the input image. Therefore, we propose the RSM to evaluate the quality of each RGB and the depth features generated from the component prototypes and adaptively weight them in proportion to their reliability. As shown in Fig.~\ref{fig:RSM} (a), the outputs of PSNM $\mathbf{\Gamma_{RGB}}$ and $\mathbf{\Gamma_D}$ processed in each encoder level are concatenated in the channel dimension for each level. These multiscale concatenated features $\mathbf{\Gamma _ {RGBD \left ( 1/8 \right ) }} \in \mathbb{R} ^ {\left( 2\times N_s \right)\times \frac{ H } { 8 } \times \frac{ W } { 8 }}$, $\mathbf{\Gamma _ {RGBD \left ( 1/16 \right ) }} \in \mathbb{R} ^ {\left( 2\times N_s \right)\times \frac{ H } { 16 } \times \frac{ W } { 16 }}$, and $\mathbf{\Gamma _ {RGBD \left ( 1/32 \right ) }} \in \mathbb{R} ^ {\left( 2\times N_s \right)\times \frac{ H } { 32 } \times \frac{ W } { 32 }}$ are then fused by applying $1\times 1$ convolution, upsampling, and element-wise summation. This fusion technique is mostly similar to FFM. The fused feature $\mathbf{F _ {RSM}} \in \mathbb{R} ^ {\left( 2\times N_s \right)\times \frac{ H } { 8 } \times \frac{ W } { 8 }}$ is then fed as input to RSM. The RSM network consists of three convolutional layers. Each layer is composed of convolution, batch normalization, and ReLU~\cite{agarap2018deep} activation. After extracting the features, we flatten the output of the last layer and apply linear function and sigmoid function. In this way, we obtain a vector $\mathbf{RelyW}\in \mathbb{R} ^ {2}$ of two reliance values $RelyW_{R}$ and $RelyW_{D}$, lying between 0 and 1, which represent the reliability of each $\mathbf{\Gamma_{RGB}}$ and $\mathbf{\Gamma_D}$ respectively. The more reliable the feature is, the higher the reliance value. Finally, we obtain a reliance-weighted RGB-D feature matrix $\mathbf{\Gamma' _ {RGBD}}$ by multiplying $RelyW_{R}$ and $RelyW_{D}$ with $\mathbf{\Gamma _ {RGBD}}$. The equations are as follows:
	
	\begin{equation}
		\mathbf{\Gamma' _ {RGBD}} = \begin{cases} RelyW_R \times \mathbf{\Gamma_{RGBD} ^ { k }} & , 0 \leq k < N_s  \\ RelyW_D \times \mathbf{\Gamma_{RGBD} ^ { k }} & , N_s \leq k < 2 \times N_s \end{cases},
	\end{equation}
	
	\noindent where $k$ indicates the channel dimension. 
	
	\noindent
	\textbf{Ground truth for RSM.} To optimize RSM, we generate a ground truth vector $\mathbf{RSM_{gt}}$. The process is demonstrated in Fig.~\ref{fig:RSM} (b). First, we process the channel-wise summation of $\mathbf{F_{RSM}}$ and apply min-max normalization. Thereby, we obtain a one-channel matrix $\mathbf{PseudoGT}\in \mathbb{R} ^ {1\times \frac{ H } { 8 } \times \frac{ W } { 8 }}$ which contains the channel-wise statistics for width $\times$ height dimensions in $\mathbf{F_{RSM}}$. Because each channel of $\mathbf{F_{RSM}}$ represents a candidate for the correlation map compressed to a small size, the bigger the pixel value is, the more likely that pixel belongs to the salient object. Next, we calculate the $L_1$ distance between $\mathbf{PseudoGT}$ and each channel of $\mathbf{F_{RSM}}$ to obtain a distance matrix $\mathbf{D_{RSM}} \in \mathbb{R} ^ {\left( 2\times N_s \right)\times \frac{ H } { 8 } \times \frac{ W } { 8 }}$. From $\mathbf{D_{RSM}}$, we obtain the mean distance values $D_R$ and $D_D$ by averaging the values where the channel index $k$ is $0 \leq k < N_s$ and $N_s \leq k < 2 \times N_s$, respectively. Finally, we acquire the two elements of $\mathbf{RSM_{gt}}$ by the following equations:
	\begin{equation}
		RSM_{{gt}_{R}} = D_D / (D_R+D_D)
	\end{equation}
	\begin{equation}
		RSM_{{gt}_{D}} = D_R / (D_R+D_D)
	\end{equation}
	Therefore, $RSM_{{gt}_{R}}$ and $RSM_{{gt}_{D}}$ represent the similarity between the RGB correlation maps and $\mathbf{PseudoGT}$, and the similarity between the depth correlation maps and $\mathbf{PseudoGT}$, respectively, which are in other words, the reliability. 
	With the generated $\mathbf{RSM_{gt}}$, RSM is optimized by minimizing the $L_1$ distance between $\mathbf{RelyW}$ and $\mathbf{RSM_{gt}}$.
	
	\subsection{Model Optimization}
	\label{Optimization}
	We optimize the model with three object functions $L_{mask}$, $L_{PSNM}$ and $L_{RSM}$. First, $L_{mask}$ is the intersection over union (IOU) loss between the predicted saliency map $I_{pred}$ and the ground truth mask $I_{GT}$, expressed as:
	
	\begin{equation}
		L _ { mask } \left ( I _ { pred } ,I _ { GT } \right ) =1- \frac{ \sum _ { x,y } ^ { } min \left ( I _ { pred \left ( x,y \right ) } ,I _ { GT \left ( x,y \right ) } \right ) } { \sum _ { x,y } ^ { } max \left ( I _ { pred \left ( x,y \right ) } ,I _ { GT \left ( x,y \right ) } \right ) },
	\end{equation}
	
	\noindent
	where $\left( x,y \right)$ are the pixel coordinates. Next, as described in section~\ref{PSNM}, $L_{PSNM}$ is the binary cross entropy loss between $\mathbf{AM_{Pred}}$ and $\mathbf{AM_{GT}}$. Finally, the loss function $L_{RSM}$ for RSM is defined by measuring the $L_1$ distance between $\mathbf{RelyW}$ and $\mathbf{RSM_{gt}}$. $L_{RSM}$ is expressed as:
	
	\begin{equation}
		L_{RSM} \left (\mathbf{RelyW}, \mathbf{RSM_{gt}} \right) = |\mathbf{RelyW}, \mathbf{RSM_{gt}}|
	\end{equation}
	
	\noindent
	As a result, we combine all these constraints regarding PSNM, RSM, and $I_{pred}$, and obtain the following objective function $L$:
	
	\begin{equation}
		\label{eqn:total}
		L_{total} = \lambda_m L_{mask} + \lambda_p L_{PSNM} + \lambda_r L_{RSM},
	\end{equation}
	where $\lambda_m$, $\lambda_p$, and $\lambda_r$ denote the weights controlling the contribution of each multiplied loss function.

	\section{Experiments}
	\subsection{Datasets}
	We perform our experiments on the following five popular RGB-D SOD benchmarks to validate the effectiveness of our proposed method: NJU2K~\cite{ju2014depth}, NLPR \cite{peng2014rgbd}, STERE~\cite{niu2012leveraging}, DES~\cite{cheng2014depth} and SIP~\cite{fan2020rethinking}. 
	NJU2K~\cite{ju2014depth} and NLPR~\cite{peng2014rgbd} consists of 1985 and 1000 paired stereoscopic images, respectively. STERE~\cite{niu2012leveraging} consists of 1000 stereo images collected from the Internet. DES~\cite{cheng2014depth}, which is also called RGBD135 in some other papers captures seven indoor scenes and contains 135 indoor images acquired by Microsoft Kinect. SIP~\cite{fan2020rethinking} is a high-quality dataset with 929 images. To make a fair comparison with previous works, we conduct experiments with two different training setups. First, we use 1485 samples from NJU2K~\cite{ju2014depth} and 700 samples from NLPR~\cite{peng2014rgbd} following the same setup as~\cite{zhao2019contrast,li2020cross,luo2020cascade}. Second, we follow the same training settings as existing works~\cite{piao2019depth,zhang2020select,ji2020accurate,chen2020progressively,zhang2020asymmetric,sun2021deep,ji2021calibrated}, using 800 samples from DUT-RGBD~\cite{piao2019depth}, 1485 samples from NJU2K~\cite{ju2014depth} and 700 samples from NLPR~\cite{peng2014rgbd}.
	
	\subsection{Evaluation Metrics}
	We evaluate the performance of our method and other methods using
	five widely used evaluation metrics: the mean F-measure ($F_{\beta}$)~\cite{achanta2009frequency}, mean absolute error (MAE, $M$)~\cite{borji2015salient}, S-measure ($S_{\alpha}$)~\cite{fan2017structure}, E-measure ($E_{\xi}$)~\cite{fan2018enhanced}, and precision-recall (PR) curve.
	
	\begin{table}[t]
		\scriptsize
		\begin{center}
			\caption{Quantitative comparison on five representative large-scale benchmark datasets. $\uparrow$ indicates that higher is better and $\downarrow$ indicates that lower is better. $*$ denotes the models are trained on NJU2K~\cite{ju2014depth} and NLPR~\cite{peng2014rgbd}; the rest are trained on DUT-RGBD~\cite{piao2019depth}, NJU2K~\cite{ju2014depth}, and NLPR~\cite{peng2014rgbd}. The best and second best are highlighted in \textcolor{red}{red} and \textcolor{blue}{blue}, respectively.}
			\label{table:qua}
			\begin{tabular}{p{0.7cm}<{\centering}|c|p{0.85cm}<{\centering}cp{0.75cm}<{\centering}p{0.75cm}<{\centering}cccccp{0.75cm}<{\centering}|cc}
				\hline
				\hline
				
				\multirow{2}{*}{} & \multirow{2}{*}{Metric} & DMRA & CPFP & CIM & CoN & CMWN & PGAR & CasG & ATS & D2F & DCF & Ours & Ours  \\
				&  & \cite{piao2019depth} & \cite{zhao2019contrast} $*$ & \cite{zhang2020select} & \cite{ji2020accurate} & \cite{li2020cross} $*$ & \cite{chen2020progressively} & \cite{luo2020cascade} $*$ & \cite{zhang2020asymmetric} & \cite{sun2021deep} & \cite{ji2021calibrated} & $*$ & \\ \hline
				\multirow{2}{*}{} & \multirow{2}{*}{Years} & ICCV & CVPR & CVPR & ECCV & ECCV & ECCV & ECCV & ECCV & CVPR & CVPR & &  \\ 
				&  & 2019 & 2019 & 2020 & 2020 & 2020 & 2020 & 2020 & 2020 & 2021 & 2021 & & \\ \hline
				
				\multirow{4}{*}{\rotatebox{90}{NJU2K}~\rotatebox{90}{~\cite{ju2014depth}}}    
				& $E_{\xi}\uparrow$ 
				& .908 & .895 & - & .912 & .936 & .916 & .877 & .921 & .923 & .922 & \textcolor{blue}{.943} & \textcolor{red}{.950} \\
				& $S_{\alpha}\uparrow$ 
				& .886 & .878 & .899 & .894 & .903 & .909 & .849 & .901 & .903 & - & \textcolor{blue}{.912} & \textcolor{red}{.918} \\
				& $F_{\beta}\uparrow$ 
				& .872 & .837 & .886 & .872 & .902 & .893 & .864 & .893 & .901 & .897 & \textcolor{blue}{.912} & \textcolor{red}{.920} \\
				& $M\,\downarrow$ 
				& .051 & .053 & .043 & .047 & .046 & .042 & .073 & .040 & .039 & .038 & \textcolor{blue}{.033} & \textcolor{red}{.032} \\ \hline
				
				\multirow{4}{*}{\rotatebox{90}{NLPR}~\rotatebox{90}{~\cite{peng2014rgbd}}}    
				& $E_{\xi}\uparrow$ 
				& .941 & .924 & - & .936 & .951 & .955 & .952 & .945 & .950 & .956 & \textcolor{red}{.962} & \textcolor{blue}{.958} \\
				& $S_{\alpha}\uparrow$ 
				& .899 & .888 & .914 & .907 & .917 & \textcolor{red}{.930} & .919 & .907 & .918 & - & \textcolor{blue}{.926} & .923 \\
				& $F_{\beta}\uparrow$ 
				& .854 & .822 & .875 & .848 & .903 & .885 & .904 & .876 & .897 & .893 & \textcolor{red}{.914} & \textcolor{blue}{.910} \\
				& $M\,\downarrow$ 
				& .031 & .036 & .026 & .031 & .029 & .024 & .025 & .028 & .024 & \textcolor{blue}{.023} & \textcolor{red}{.022} & \textcolor{blue}{.023} \\ \hline
				
				\multirow{4}{*}{\rotatebox{90}{STERE}~\rotatebox{90}{~~\cite{niu2012leveraging}}}    
				& $E_{\xi}\uparrow$ 
				& .920 & .903 & - & .923 & \textcolor{red}{.944} & .919 & .930 & .921 & .933 & .931 & .942 & \textcolor{blue}{.943} \\
				& $S_{\alpha}\uparrow$ 
				& .886 & .879 & .893 & \textcolor{red}{.908} & .905 & .913 & .899 & .897 & .904 & - & .906 & \textcolor{blue}{.907} \\
				& $F_{\beta}\uparrow$ 
				& .867 & .830 & .880 & .885 & \textcolor{red}{.901} & .880 & \textcolor{red}{.901} & .884 & .898 & .890 & .898 & \textcolor{blue}{.900} \\
				& $M\,\downarrow$ 
				& .047 & .051 & .044 & .041 & .043 & .041 & .039 & .039 & \textcolor{blue}{.036} & .037 & \textcolor{red}{.035} & \textcolor{red}{.035} \\ \hline
				
				\multirow{4}{*}{\rotatebox{90}{DES}~\rotatebox{90}{~\cite{cheng2014depth}}}    
				& $E_{\xi}\uparrow$ 
				& .944 & .927 & - & .945 & .969 & .939 & .947 & .952 & .962 & - & \textcolor{red}{.976} & \textcolor{blue}{.974} \\
				& $S_{\alpha}\uparrow$ 
				& .900 & .872 & .905 & .910 & .934 & .913 & .905 & .907 & .920 & - & \textcolor{red}{.938} & \textcolor{blue}{.937} \\
				& $F_{\beta}\uparrow$ 
				& .866 & .829 & .876 & .861 & .930 & .880 & .906 & .885 & .896 & - & \textcolor{red}{.943} & \textcolor{blue}{.936} \\
				& $M\,\downarrow$ 
				& .030 & .038 & .025 & .027 & .022 & .026 & .028 & .024 & \textcolor{blue}{.021} & - & \textcolor{red}{.016} & \textcolor{red}{.016} \\ \hline
				
				\multirow{4}{*}{\rotatebox{90}{SIP}~\rotatebox{90}{~\cite{fan2020rethinking}}}    
				& $E_{\xi}\uparrow$ 
				& .863 & .899 & - & .909 & .913 & .908 & - & - & - & .920 & \textcolor{red}{.936} & \textcolor{blue}{.934} \\
				& $S_{\alpha}\uparrow$ 
				& .806 & .850 & - & .858 & .867 & .876 & - & - & - & - & \textcolor{blue}{.890} & \textcolor{red}{.892} \\
				& $F_{\beta}\uparrow$ 
				& .819 & .819 & - & .842 & .874 & .854 & - & - & - & .877 & \textcolor{blue}{.896} & \textcolor{red}{.899} \\
				& $M\,\downarrow$ 
				& .085 & .064 & - & .063 & .062 & .055 & - & - & - & \textcolor{blue}{.051} & \textcolor{red}{.042} & \textcolor{red}{.042} \\ \hline
				\hline
			\end{tabular}
		\end{center}
	\end{table}
	
	\begin{figure}[t!]
		\setlength{\belowcaptionskip}{-24pt}
		\begin{center}
			\includegraphics[width=1\linewidth]{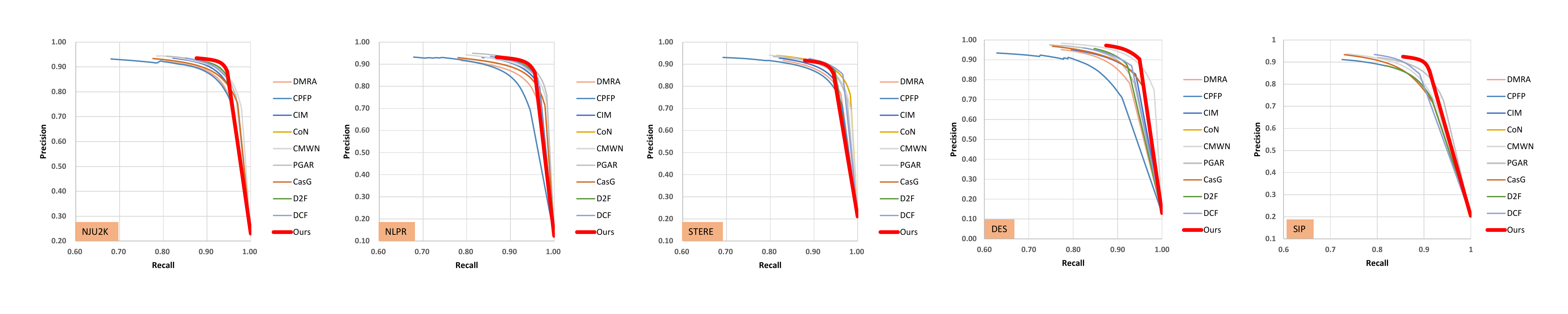}
			\caption{Precision-recall curve comparison on five datasets.}
			\label{fig:PR}
		\end{center}
	\end{figure}

	\subsection{Implementation Details}
	We set the number of superpixels $N_S$ to 100 and $a_k$ of EdgeConv to 10. We also set $\lambda_m$, $\lambda_p$, and $\lambda_r$ in Equation~\ref{eqn:total} to 1, 1, and 10, respectively, for balanced training. We implement the proposed method using the open deep learning framework PyTorch. The backbone network is equipped with VGG-16~\cite{simonyan2014very}, with initial parameters pre-trained in ImageNet~\cite{deng2009imagenet}. All images are uniformly resized to $352 \times 352$ pixels for training and inferring. For network training, we used the Adam optimizer~\cite{kingma2014adam} with $\beta_1=0.9$, $\beta_2=0.999$, and $\epsilon =10 ^ { -8 }$. The learning rate decayed from $8 \times 10^{-5}$ to $8 \times 10^{-6}$ with the cosine annealing scheduler~\cite{loshchilov2016sgdr}. The total number of epochs was set to 200 with batch size 16 with two NVIDIA RTX 3090 GPUs for all experiments in this study.
	
	\begin{figure}[t]
		\setlength{\belowcaptionskip}{-24pt}
		\begin{center}
			\includegraphics[width=\linewidth]{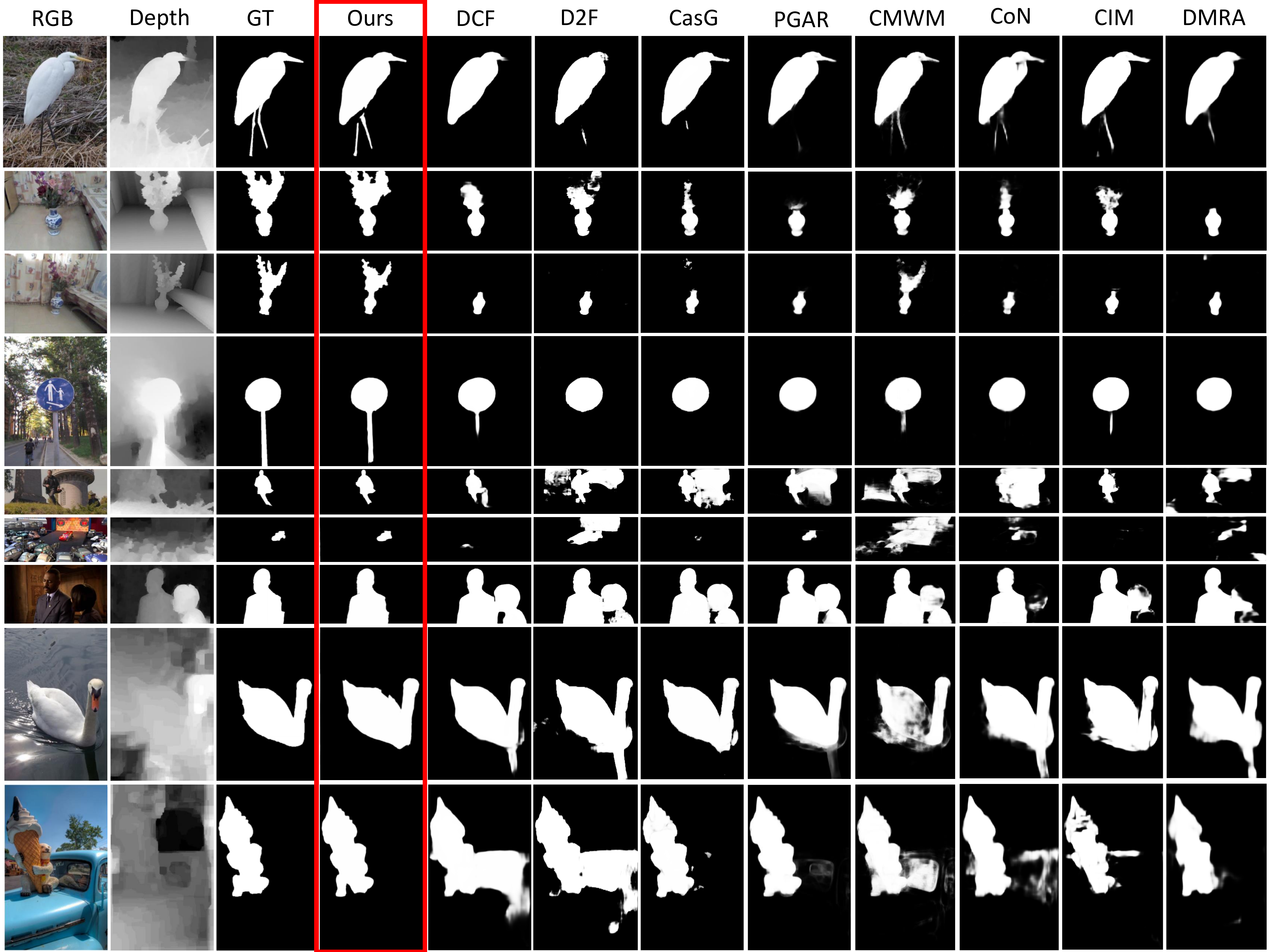}
			\caption{Qualitative comparison with eight state-of-the-art methods.}
			\label{fig7}
		\end{center}
	\end{figure} 
	
	\subsection{Comparison with State-of-the-Art Methods}
	\textbf{Quantitative comparison.} Table~\ref{table:qua} shows our quantitative performance compared with 10 recently published state-of-the-art RGB-D SOD methods, DMRA \cite{piao2019depth}, CPFP~\cite{zhao2019contrast}, CIM~\cite{zhang2020select}, CoN~\cite{ji2020accurate}, CMWN~\cite{li2020cross}, GAR~\cite{chen2020progressively}, CasG~\cite{luo2020cascade}, ATS~\cite{zhang2020asymmetric}, D2F~\cite{sun2021deep}, DCF~\cite{ji2021calibrated}, on five popular benchmark datasets. Because the training data in these comparative studies differ slightly, as some used NJU2K~\cite{ju2014depth} and NLPR~\cite{peng2014rgbd} whereas others also used DUT-RGBD~\cite{piao2019depth}, we show the performance on both settings for a fair comparison. It is observed that our model notably outperforms the other methods. In particular, our model exceeds the counterpart methods by a dramatic margin in terms of all four evaluation metrics on NJU2K~\cite{ju2014depth} and DES~\cite{cheng2014depth}, which are considered more challenging than to the others due to the low contrast and objects cluttering the background. This result further indicates that our network can perform well on various complex scenes. Moreover, we plotted the PR curves in Fig.~\ref{fig:PR} for a better comparison. The results show that ours lies above most of the methods compared.
	
	\noindent \textbf{Qualitative comparison.} In Fig.~\ref{fig7}, we compare our qualitative results to those of eight top-ranking RGB-D SOD approaches on several challenging scenarios, including low contrast, reflection, thin objects, multiple objects, and long distance. Particularly for scenes with complex RGB maps caused by cluttered objects and patterns in the background (e.g., the second, third, and fourth row), our model utilized more information from the reliable depth maps to generate an accurate saliency map. Furthermore, the accuracy of such scenes is boosted by the PSNM, which effectively discriminates the foreground from the background. Similarly, our model can handle samples with depth maps that are ambiguous because of light reflection and long distance (e.g., the sixth and eighth rows) because our model adaptively decides to rely on the more accurate RGB maps. Furthermore, it is observed that our model is robust to scenes with multiple objects (e.g., the fifth and ninth rows).
	
	\begin{table}[t]
		\caption{Performance with different combinations of our contributions. RE and DE represent the encoders for the RGB and depth, respectively. RS and DS are the set of FFM, PGM, and PSNM in the RGB and depth streams, respectively.}
		\label{table:ab1}
		\resizebox{1\columnwidth}{!}{
			\begin{tabular}{c|ccccc|cccc|cccc|cccc|cccc|cccc}
				\hline
				\multirow{2}{*}{Index} & \multicolumn{5}{c|}{Method} & \multicolumn{4}{c|}{NJU2K~\cite{ju2014depth}}    & \multicolumn{4}{c|}{NLPR~\cite{peng2014rgbd}}     & \multicolumn{4}{c|}{STERE~\cite{niu2012leveraging}}    & \multicolumn{4}{c|}{DES~\cite{cheng2014depth}}      & \multicolumn{4}{c}{SIP~\cite{fan2020rethinking}}       \\ \cline{2-26} 
				& RE  & DE  & RS  & DS  & RSM & $E_{\xi}\uparrow$    & $S_{\alpha}\uparrow$    & $F_{\beta}\uparrow$    & $M\downarrow$    & $E_{\xi}\uparrow$    & $S_{\alpha}\uparrow$    & $F_{\beta}\uparrow$    & $M\downarrow$    & $E_{\xi}\uparrow$    & $S_{\alpha}\uparrow$    & $F_{\beta}\uparrow$    & $M\downarrow$    & $E_{\xi}\uparrow$    & $S_{\alpha}\uparrow$    & $F_{\beta}\uparrow$    & $M\downarrow$    & $E_{\xi}\uparrow$    & $S_{\alpha}\uparrow$    & $F_{\beta}\uparrow$    & $M\downarrow$ \\ \hline
				(a)                    & \ding{51}  &     &     &     &     & .872 & .836 & .848 & .058 & .871 & .852 & .793 & .044 & .883 & .852 & .843 & .061 & .843 & .775 & .764 & .048 & .825 & .738 & .734 & .069 \\
				(b)                    & \ding{51}  & \ding{51}  &     &     &     & .904 & .863 & .869 & .051 & .912 & .877 & .842 & .037 & .908 & .870 & .864 & .052 & .888 & .831 & .820 & .038 & .868 & .796 & .793 & .062 \\
				(c)                    & \ding{51}  & \ding{51}  &     &     & \ding{51}  & .934 & .888 & .887 & .044 & .952 & .905 & .888 & .028 & .934 & .889 & .883 & .042 & .932 & .887 & .877 & .031 & .913 & .855 & .853 & .057 \\
				(d)                    & \ding{51}  &     & \ding{51}  &     &     & .937 & .903 & .901 & .039 & .950 & .915 & .896 & .026 & .938 & .902 & .894 & .037 & .950 & .905 & .900 & .027 & .918 & .870 & .878 & .051 \\
				(e)                    & \ding{51}  & \ding{51}  & \ding{51}  & \ding{51}  &     & .944 & .912 & .910 & .035 & .954 & .919 & .902 & .025 & .940 & .905 & .898 & .036 & .963 & .922 & .917 & .022 & .926 & .881 & .887 & .047 \\
				(f)                    & \ding{51}  & \ding{51}  & \ding{51}  & \ding{51}  & \ding{51}  & \bf{.950} & \bf{.918} & \bf{.920} & \bf{.032} & \bf{.958} & \bf{.923} & \bf{.910} & \bf{.023} & \bf{.943} & \bf{.907} & \bf{.900} & \bf{.035} & \bf{.974} & \bf{.937} & \bf{.936} & \bf{.016} & \bf{.934} & \bf{.892} & \bf{.899} & \bf{.042} \\ \hline
			\end{tabular}
		}
	\end{table}
	
	\begin{figure}[!t]
		\setlength{\belowcaptionskip}{-24pt}
		\begin{center}
			\includegraphics[width=0.9\linewidth]{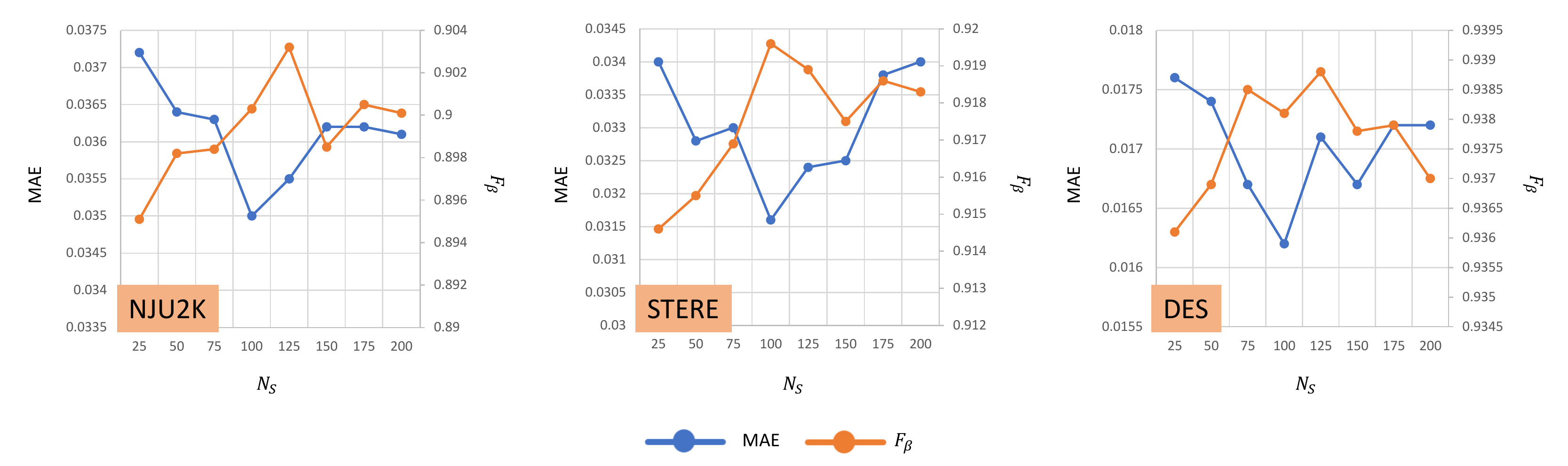}
			\caption{Comparison of performance characteristics with respect to $N_S$ for the NJU2K~\cite{ju2014depth}, STERE~\cite{niu2012leveraging}, and DES~\cite{cheng2014depth} datasets.}
			\label{fig:ab2}
		\end{center}
	\end{figure}
	
	\subsection{Ablation Analysis}
	We verify the performance of our model through various ablation studies. Table~\ref{table:ab1} shows the effects of the proposed modules in various combinations. RE and DE in Table~\ref{table:ab1} represent the VGG-16~\cite{simonyan2014very} encoders for the RGB images and depth maps, respectively. In addition, RS and DS are the set of FFM, PGM, and PSNM in the RGB and depth streams, respectively. The proposed RSM only applies when both RE and DE are used. Fig~\ref{fig:ab2} also shows the performance of our model according to the number of superpixels.
	
\begin{figure}[t]
		\includegraphics[width=\linewidth]{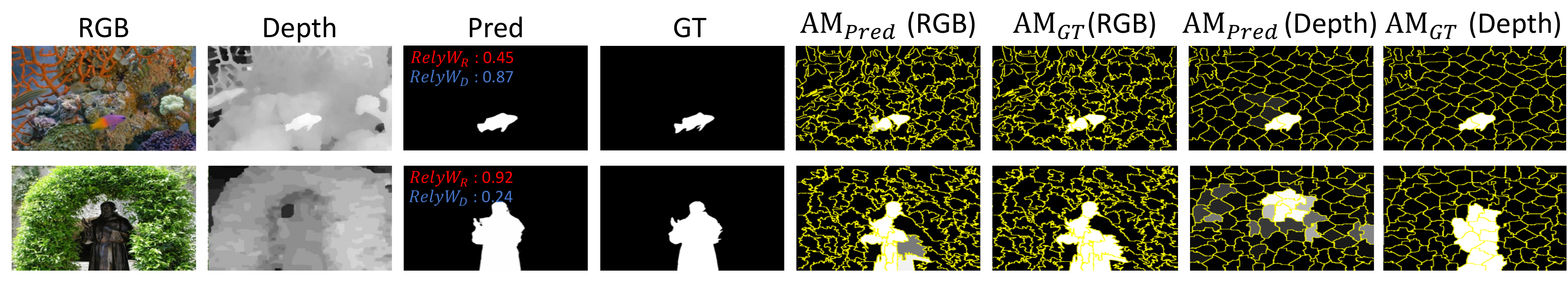}
		\caption{Visualization of our results in several challenging situations. $AM_{pred}$ and $AM_{GT}$ are described in Section 3.4, and $RelyW_{R}$ and $RelyW_{D}$ are described in Section~\ref{RSM}}
		\label{fig9}
	\end{figure}

	\noindent
	\textbf{Impact of prototype sampling.} As shown in Table~\ref{table:ab1}, (d) and (e), to which the prototype sampling method is applied, our method achieves better performance than (a) and (b) on all datasets. This is because the encoder-decoder-based network delivers not only the features for the salient object but also the background and non-salient object feature information extracted from the encoder to the decoder, preventing accurate mask generation. In contrast, the proposed SPSN model performs well because the network can selectively extract only the important salient object feature information by PSNM. Furthermore, Fig.~\ref{fig9} shows the salient prototype sampling results of the proposed method.
	
	\noindent
	\textbf{Impact of RSM.} When the proposed RSM module is applied, as shown in Table~\ref{table:ab1} (c) and (f), it shows significant performance improvement when the RGB image and depth map are used together. The performance improves because RSM selects feature maps generated from RGB and depth streams based on their reliability. Therefore, as shown in Fig.~\ref{fig9}, $RelyW_{R}$ is small for RGB images with camouflaged objects, and $RelyW_{D}$ is small for low-quality depth maps. This structure shows that the model reduces the biased dependence and makes it robust to low-quality depth maps.
	
	\noindent
	\textbf{Number of superpixels.} We conduct ablation studies to observe how the MAE and $F_{\beta}$ values change according to the number of superpixels $N_S$. Fig.~\ref{fig:ab2} shows the changes in performance using the NJU2K~\cite{ju2014depth}, STERE~\cite{niu2012leveraging}, and DES~\cite{cheng2014depth} datasets according to the number of superpixels. As shown in Fig.~\ref{fig:ab2}, the proposed model performs best near $N_S = 100$. Additionally, if $N_S$ is too small or too large, the performance will decrease. This degraded performance results because if $N_S$ is too small, the superpixel masks cannot effectively separate salient and non-salient objects and cannot provide a sufficient number of component prototypes. Conversely, if $N_S$ is too large, it is difficult to create coherent features for the salient object by creating too small superpixel masks that are too small.
	
	\section{Conclusion}
	In this paper, we aim to segment salient objects by designing an SPSN, which suppresses the effects of background objects and effectively takes advantage of RGB and depth maps. Specifically, our network is composed of four novel modules—the FFM, which fuses the multiscale features extracted from the encoder; the PGM, which renders the fused feature maps to component prototypes; the PSNM, which discriminates the prototype that belongs to the salient object; and the RSM, which adaptively selects the contribution of RGB and depth features. The results demonstrate the outstanding improvement of our method over the previous studies, indicating that our model can capture salient objects in various challenging scenes. Furthermore, extensive ablation studies show the contribution and effectiveness of each of the proposed modules.
	
	\vfill
	\noindent\footnotesize\textbf{Acknowledgement.} This work was supported by the Institute of Information \& communications Technology Planning \& Evaluation(IITP) grant funded by the Korea government(MSIT) (No. 2021-0-00172, The development of human Re-identification and masked face recognition based on CCTV camera).
	\pagebreak
	
	\clearpage
	\normalsize
	\bibliographystyle{splncs04}
	\bibliography{egbib}

\begin{thebibliography}{10}
\providecommand{\url}[1]{\texttt{#1}}
\providecommand{\urlprefix}{URL }
\providecommand{\doi}[1]{https://doi.org/#1}

\bibitem{achanta2009frequency}
Achanta, R., Hemami, S., Estrada, F., Susstrunk, S.: Frequency-tuned salient
  region detection. In: 2009 IEEE conference on computer vision and pattern
  recognition. pp. 1597--1604. IEEE (2009)

\bibitem{achanta2012slic}
Achanta, R., Shaji, A., Smith, K., Lucchi, A., Fua, P., S{\"u}sstrunk, S.: Slic
  superpixels compared to state-of-the-art superpixel methods. IEEE
  transactions on pattern analysis and machine intelligence  \textbf{34}(11),
  2274--2282 (2012)

\bibitem{agarap2018deep}
Agarap, A.F.: Deep learning using rectified linear units (relu). arXiv preprint
  arXiv:1803.08375  (2018)

\bibitem{borji2015salient}
Borji, A., Cheng, M.M., Jiang, H., Li, J.: Salient object detection: A
  benchmark. IEEE transactions on image processing  \textbf{24}(12),
  5706--5722 (2015)

\bibitem{chen2017m}
Chen, H., Li, Y.F., Su, D.: M 3 net: Multi-scale multi-path multi-modal fusion
  network and example application to rgb-d salient object detection. In: 2017
  IEEE/RSJ International Conference on Intelligent Robots and Systems (IROS).
  pp. 4911--4916. IEEE (2017)

\bibitem{chen2018attention}
Chen, H., Li, Y.F., Su, D.: Attention-aware cross-modal cross-level fusion
  network for rgb-d salient object detection. In: 2018 IEEE/RSJ International
  Conference on Intelligent Robots and Systems (IROS). pp. 6821--6826. IEEE
  (2018)

\bibitem{chen2018progressively}
Chen, H., Li, Y.: Progressively complementarity-aware fusion network for rgb-d
  salient object detection. In: Proceedings of the IEEE conference on computer
  vision and pattern recognition. pp. 3051--3060 (2018)

\bibitem{chen2019three}
Chen, H., Li, Y.: Three-stream attention-aware network for rgb-d salient object
  detection. IEEE Transactions on Image Processing  \textbf{28}(6),  2825--2835
  (2019)

\bibitem{chen2017deeplab}
Chen, L.C., Papandreou, G., Kokkinos, I., Murphy, K., Yuille, A.L.: Deeplab:
  Semantic image segmentation with deep convolutional nets, atrous convolution,
  and fully connected crfs. IEEE transactions on pattern analysis and machine
  intelligence  \textbf{40}(4),  834--848 (2017)

\bibitem{chen2020progressively}
Chen, S., Fu, Y.: Progressively guided alternate refinement network for rgb-d
  salient object detection. In: European Conference on Computer Vision. pp.
  520--538. Springer (2020)

\bibitem{cheng2014depth}
Cheng, Y., Fu, H., Wei, X., Xiao, J., Cao, X.: Depth enhanced saliency
  detection method. In: Proceedings of international conference on internet
  multimedia computing and service. pp. 23--27 (2014)

\bibitem{cong2019going}
Cong, R., Lei, J., Fu, H., Hou, J., Huang, Q., Kwong, S.: Going from rgb to
  rgbd saliency: A depth-guided transformation model. IEEE transactions on
  cybernetics  \textbf{50}(8),  3627--3639 (2019)

\bibitem{cong2016saliency}
Cong, R., Lei, J., Zhang, C., Huang, Q., Cao, X., Hou, C.: Saliency detection
  for stereoscopic images based on depth confidence analysis and multiple cues
  fusion. IEEE Signal Processing Letters  \textbf{23}(6),  819--823 (2016)

\bibitem{deng2009imagenet}
Deng, J., Dong, W., Socher, R., Li, L.J., Li, K., Fei-Fei, L.: Imagenet: A
  large-scale hierarchical image database. In: 2009 IEEE conference on computer
  vision and pattern recognition. pp. 248--255. Ieee (2009)

\bibitem{desingh2013depth}
Desingh, K., Krishna, K.M., Rajan, D., Jawahar, C.: Depth really matters:
  Improving visual salient region detection with depth. In: BMVC. pp. 1--11
  (2013)

\bibitem{fan2018salient}
Fan, D.P., Cheng, M.M., Liu, J.J., Gao, S.H., Hou, Q., Borji, A.: Salient
  objects in clutter: Bringing salient object detection to the foreground. In:
  Proceedings of the European conference on computer vision (ECCV). pp.
  186--202 (2018)

\bibitem{fan2017structure}
Fan, D.P., Cheng, M.M., Liu, Y., Li, T., Borji, A.: Structure-measure: A new
  way to evaluate foreground maps. In: Proceedings of the IEEE international
  conference on computer vision. pp. 4548--4557 (2017)

\bibitem{fan2018enhanced}
Fan, D.P., Gong, C., Cao, Y., Ren, B., Cheng, M.M., Borji, A.:
  Enhanced-alignment measure for binary foreground map evaluation. arXiv
  preprint arXiv:1805.10421  (2018)

\bibitem{fan2020rethinking}
Fan, D.P., Lin, Z., Zhang, Z., Zhu, M., Cheng, M.M.: Rethinking rgb-d salient
  object detection: Models, data sets, and large-scale benchmarks. IEEE
  Transactions on neural networks and learning systems  \textbf{32}(5),
  2075--2089 (2020)

\bibitem{fu2019dual}
Fu, J., Liu, J., Tian, H., Li, Y., Bao, Y., Fang, Z., Lu, H.: Dual attention
  network for scene segmentation. In: Proceedings of the IEEE/CVF conference on
  computer vision and pattern recognition. pp. 3146--3154 (2019)

\bibitem{fu2020jl}
Fu, K., Fan, D.P., Ji, G.P., Zhao, Q.: Jl-dcf: Joint learning and
  densely-cooperative fusion framework for rgb-d salient object detection. In:
  Proceedings of the IEEE/CVF conference on computer vision and pattern
  recognition. pp. 3052--3062 (2020)

\bibitem{guo2016salient}
Guo, J., Ren, T., Bei, J.: Salient object detection for rgb-d image via
  saliency evolution. In: 2016 IEEE International Conference on Multimedia and
  Expo (ICME). pp.~1--6. IEEE (2016)

\bibitem{ji2021calibrated}
Ji, W., Li, J., Yu, S., Zhang, M., Piao, Y., Yao, S., Bi, Q., Ma, K., Zheng,
  Y., Lu, H., et~al.: Calibrated rgb-d salient object detection. In:
  Proceedings of the IEEE/CVF Conference on Computer Vision and Pattern
  Recognition. pp. 9471--9481 (2021)

\bibitem{ji2020accurate}
Ji, W., Li, J., Zhang, M., Piao, Y., Lu, H.: Accurate rgb-d salient object
  detection via collaborative learning. In: Computer Vision--ECCV 2020: 16th
  European Conference, Glasgow, UK, August 23--28, 2020, Proceedings, Part
  XVIII 16. pp. 52--69. Springer (2020)

\bibitem{ju2014depth}
Ju, R., Ge, L., Geng, W., Ren, T., Wu, G.: Depth saliency based on anisotropic
  center-surround difference. In: 2014 IEEE international conference on image
  processing (ICIP). pp. 1115--1119. IEEE (2014)

\bibitem{kingma2014adam}
Kingma, D.P., Ba, J.: Adam: A method for stochastic optimization. arXiv
  preprint arXiv:1412.6980  (2014)

\bibitem{lang2012depth}
Lang, C., Nguyen, T.V., Katti, H., Yadati, K., Kankanhalli, M., Yan, S.: Depth
  matters: Influence of depth cues on visual saliency. In: European conference
  on computer vision. pp. 101--115. Springer (2012)

\bibitem{li2021adaptive}
Li, G., Jampani, V., Sevilla-Lara, L., Sun, D., Kim, J., Kim, J.: Adaptive
  prototype learning and allocation for few-shot segmentation. In: Proceedings
  of the IEEE/CVF Conference on Computer Vision and Pattern Recognition. pp.
  8334--8343 (2021)

\bibitem{li2020cross}
Li, G., Liu, Z., Ye, L., Wang, Y., Ling, H.: Cross-modal weighting network for
  rgb-d salient object detection. In: European Conference on Computer Vision.
  pp. 665--681. Springer (2020)

\bibitem{liu2020part}
Liu, Y., Zhang, X., Zhang, S., He, X.: Part-aware prototype network for
  few-shot semantic segmentation. In: European Conference on Computer Vision.
  pp. 142--158. Springer (2020)

\bibitem{loshchilov2016sgdr}
Loshchilov, I., Hutter, F.: Sgdr: Stochastic gradient descent with warm
  restarts. arXiv preprint arXiv:1608.03983  (2016)

\bibitem{luo2020cascade}
Luo, A., Li, X., Yang, F., Jiao, Z., Cheng, H., Lyu, S.: Cascade graph neural
  networks for rgb-d salient object detection. In: European Conference on
  Computer Vision. pp. 346--364. Springer (2020)

\bibitem{niu2012leveraging}
Niu, Y., Geng, Y., Li, X., Liu, F.: Leveraging stereopsis for saliency
  analysis. In: 2012 IEEE Conference on Computer Vision and Pattern
  Recognition. pp. 454--461. IEEE (2012)

\bibitem{park2021saliency}
Park, C., Lee, M., Cho, M., Lee, S.: Saliency detection via global context
  enhanced feature fusion and edge weighted loss. arXiv preprint
  arXiv:2110.06550  (2021)

\bibitem{peng2014rgbd}
Peng, H., Li, B., Xiong, W., Hu, W., Ji, R.: Rgbd salient object detection: a
  benchmark and algorithms. In: European conference on computer vision. pp.
  92--109. Springer (2014)

\bibitem{piao2019depth}
Piao, Y., Ji, W., Li, J., Zhang, M., Lu, H.: Depth-induced multi-scale
  recurrent attention network for saliency detection. In: Proceedings of the
  IEEE/CVF International Conference on Computer Vision. pp. 7254--7263 (2019)

\bibitem{qi2017pointnet}
Qi, C.R., Su, H., Mo, K., Guibas, L.J.: Pointnet: Deep learning on point sets
  for 3d classification and segmentation. In: Proceedings of the IEEE
  conference on computer vision and pattern recognition. pp. 652--660 (2017)

\bibitem{qi2017pointnet++}
Qi, C.R., Yi, L., Su, H., Guibas, L.J.: Pointnet++: Deep hierarchical feature
  learning on point sets in a metric space. Advances in neural information
  processing systems  \textbf{30} (2017)

\bibitem{qu2017rgbd}
Qu, L., He, S., Zhang, J., Tian, J., Tang, Y., Yang, Q.: Rgbd salient object
  detection via deep fusion. IEEE Transactions on Image Processing
  \textbf{26}(5),  2274--2285 (2017)

\bibitem{simonyan2014very}
Simonyan, K., Zisserman, A.: Very deep convolutional networks for large-scale
  image recognition. arXiv preprint arXiv:1409.1556  (2014)

\bibitem{song2017depth}
Song, H., Liu, Z., Du, H., Sun, G., Le~Meur, O., Ren, T.: Depth-aware salient
  object detection and segmentation via multiscale discriminative saliency
  fusion and bootstrap learning. IEEE Transactions on Image Processing
  \textbf{26}(9),  4204--4216 (2017)

\bibitem{sun2021deep}
Sun, P., Zhang, W., Wang, H., Li, S., Li, X.: Deep rgb-d saliency detection
  with depth-sensitive attention and automatic multi-modal fusion. In:
  Proceedings of the IEEE/CVF Conference on Computer Vision and Pattern
  Recognition. pp. 1407--1417 (2021)

\bibitem{wang2019panet}
Wang, K., Liew, J.H., Zou, Y., Zhou, D., Feng, J.: Panet: Few-shot image
  semantic segmentation with prototype alignment. In: Proceedings of the
  IEEE/CVF International Conference on Computer Vision. pp. 9197--9206 (2019)

\bibitem{wang2019salient}
Wang, W., Zhao, S., Shen, J., Hoi, S.C., Borji, A.: Salient object detection
  with pyramid attention and salient edges. In: Proceedings of the IEEE/CVF
  Conference on Computer Vision and Pattern Recognition. pp. 1448--1457 (2019)

\bibitem{wang2018non}
Wang, X., Girshick, R., Gupta, A., He, K.: Non-local neural networks. In:
  Proceedings of the IEEE conference on computer vision and pattern
  recognition. pp. 7794--7803 (2018)

\bibitem{wang2019dynamic}
Wang, Y., Sun, Y., Liu, Z., Sarma, S.E., Bronstein, M.M., Solomon, J.M.:
  Dynamic graph cnn for learning on point clouds. Acm Transactions On Graphics
  (tog)  \textbf{38}(5),  1--12 (2019)

\bibitem{zhang2019self}
Zhang, H., Goodfellow, I., Metaxas, D., Odena, A.: Self-attention generative
  adversarial networks. In: International conference on machine learning. pp.
  7354--7363. PMLR (2019)

\bibitem{zhang2020uc}
Zhang, J., Fan, D.P., Dai, Y., Anwar, S., Saleh, F.S., Zhang, T., Barnes, N.:
  Uc-net: Uncertainty inspired rgb-d saliency detection via conditional
  variational autoencoders. In: Proceedings of the IEEE/CVF conference on
  computer vision and pattern recognition. pp. 8582--8591 (2020)

\bibitem{zhang2020asymmetric}
Zhang, M., Fei, S.X., Liu, J., Xu, S., Piao, Y., Lu, H.: Asymmetric two-stream
  architecture for accurate rgb-d saliency detection. In: European Conference
  on Computer Vision. pp. 374--390. Springer (2020)

\bibitem{zhang2020select}
Zhang, M., Ren, W., Piao, Y., Rong, Z., Lu, H.: Select, supplement and focus
  for rgb-d saliency detection. In: Proceedings of the IEEE/CVF Conference on
  Computer Vision and Pattern Recognition. pp. 3472--3481 (2020)

\bibitem{zhao2019contrast}
Zhao, J.X., Cao, Y., Fan, D.P., Cheng, M.M., Li, X.Y., Zhang, L.: Contrast
  prior and fluid pyramid integration for rgbd salient object detection. In:
  Proceedings of the IEEE/CVF Conference on Computer Vision and Pattern
  Recognition. pp. 3927--3936 (2019)

\bibitem{zhao2019egnet}
Zhao, J.X., Liu, J.J., Fan, D.P., Cao, Y., Yang, J., Cheng, M.M.: Egnet: Edge
  guidance network for salient object detection. In: Proceedings of the
  IEEE/CVF International Conference on Computer Vision. pp. 8779--8788 (2019)

\bibitem{zhao2015saliency}
Zhao, R., Ouyang, W., Li, H., Wang, X.: Saliency detection by multi-context
  deep learning. In: Proceedings of the IEEE conference on computer vision and
  pattern recognition. pp. 1265--1274 (2015)

\end{thebibliography}
\end{document}